\documentclass[10pt,twocolumn,letterpaper]{article}

\usepackage{cvpr}
\usepackage{times}
\usepackage{epsfig}
\usepackage{graphicx}
\usepackage{amsmath}
\usepackage{amssymb}
\usepackage{multirow}
\usepackage{array}
\usepackage[ruled,vlined]{algorithm2e}
\usepackage{enumerate}
\usepackage{caption}
\usepackage[position=top]{subfig}
\usepackage{adjustbox} 
\usepackage{booktabs}
\usepackage{multirow}
\usepackage{siunitx}
\usepackage{lipsum}
\usepackage{booktabs}
\usepackage{eso-pic}

\AddToShipoutPicture*{\footnotesize \sffamily\raisebox{1.6cm}[0pt][0pt]{
    {\hspace{1.4cm}
    \raisebox{0cm}{© 2021 IEEE. Personal use of this material is permitted. Permission from IEEE must be
    obtained for all other uses, in any current or future media,}%
    \hspace{-18.1cm}
    \raisebox{-0.4cm}{including
   reprinting/republishing this material for advertising or promotional purposes, creating new collective works, for resale or redistribution to}
    \hspace{-18.1cm}
    \raisebox{-0.8cm}{servers or lists, or reuse of any copyrighted component of this work in other works.}%
  }
} }

\newcommand\blfootnote[1]{%
  \begingroup
  \renewcommand\thefootnote{}\footnote{#1}%
  \addtocounter{footnote}{-1}%
  \endgroup
}

\usepackage[pagebackref=true,breaklinks=true,letterpaper=true,colorlinks,bookmarks=false]{hyperref}

\cvprfinalcopy 

\begin{document}
\pagestyle{empty} 



\title{DFM: A Performance Baseline for Deep Feature Matching}

\author{Ufuk Efe, Kutalmis Gokalp Ince, A. Aydin Alatan\\
Department of Electrical and Electronics Engineering, Center for Image Analysis (OGAM)\\
Middle East Technical University, Ankara, Turkey\\
{\tt\small ufuk.efe, kutalmis, alatan @ metu.edu.tr}
}

\maketitle
\thispagestyle{empty}

\begin{abstract} 
A novel image matching method is proposed that utilizes learned features extracted by an off-the-shelf deep neural network to obtain a promising performance. The proposed method uses pre-trained VGG architecture as a feature extractor and does not require any additional training specific to improve matching. Inspired by well-established concepts in the psychology area, such as the Mental Rotation paradigm, an initial warping is performed as a result of a preliminary geometric transformation estimate. These estimates are simply based on dense matching of nearest neighbors at the terminal layer of VGG network outputs of the images to be matched. After this initial alignment, the same approach is repeated again between reference and aligned images in a hierarchical manner to reach a good localization and matching performance. Our algorithm achieves 0.57 and 0.80 overall scores in terms of Mean Matching Accuracy (MMA) for 1 pixel and 2 pixels thresholds respectively on Hpatches dataset \cite{balntas2017hpatches}, which indicates a better performance than the state-of-the-art. 
\end{abstract}


\section{Introduction}

Determining point correspondences between images is one of the vital and well-studied topics in computer vision. These correspondences are crucial for several applications, such as Simultaneous Localization and Mapping (SLAM), Structure-from-Motion (SfM), pose estimation, image retrieval, and image matching. 

\begin{figure}[ht]
\centering
\captionsetup{belowskip=-10pt}
\includegraphics[width=0.46\textwidth]{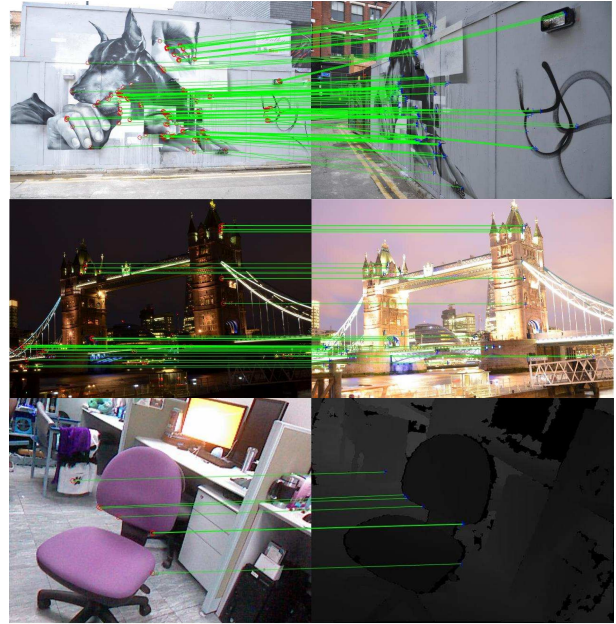}
\caption{\textbf{Visual results of the proposed method.} Proposed DFM can find high-quality matches under severe viewpoint ($1^{st}$ row) \cite{balntas2017hpatches} and illumination changes ($2^{nd}$ row) \cite{balntas2017hpatches}. It is also successful on cross-domain problems, such as RGB to depth image matching ($3^{rd}$ row) \cite{shen2014multimodal}.}
\label{fig:href}
\end{figure}

Classical image matching pipeline consists of \textit{Feature Detection, Feature Description, Feature Matching,} and \textit{Geometric Transformation Estimation} steps. Most of the popular algorithms before the deep learning era have focused on one or more phases in this pipeline. While many techniques \cite{harris1997harris,smith1997susan,rosten2005fast,matas2004mser} concentrate on Feature Detection, some other approaches \cite{dalal2005hog,mikolajczyk2005gloh,calonder2010brief} handle Feature Description step locally. The most popular algorithms \cite{lowe2004sift,bay2006surf,rublee2011orb,leutenegger2011brisk,alcantarilla2012kaze,alcantarilla2011akaze} work on both Feature Detection and Description. Feature Matching is usually achieved by minimizing/maximizing some well-defined metrics, such as Sum of Squared Differences (SSD) or correlation,  which is next accompanied by outlier rejection methods, such as ratio test of SIFT \cite{lowe2004sift} or some outlier rejection algorithms, such as LPM \cite{ma2019lpm}. In Geometric Transformation Estimation stage, either epipolar geometry or homography is estimated by DLT \cite{multiplevg} or RANSAC \cite{fischler1981ransac} based algorithms \cite{chum2005prosac,torr2000mlesac,chum2003lo-ransac,chum2005degensac} commonly. 

\blfootnote{This research is funded by ROKETSAN.}
During recent years, image matching met deep learning, and a number of valuable efforts have been performed. However, most of these studies still rely on the classical pipeline. In this respect, while many methods \cite{verdie2015tilde,lenc2016learning,savinov2017quad,detone2017magicpoint,mishkin2018repeatability,barroso2019key.net} try to improve Feature Detection with deep architectures, some others \cite{zagoruyko2015learning,simo2015discriminative,balntas2016pn-net,mishchuk2017hardnet,tian2019sosnet} reconstruct Feature Descriptors by utilizing learned representations. On the other hand, there are also techniques \cite{detone2018superpoint,ono2018lfnet,shen2019rfnet,revaud2019r2d2,dusmanu2019d2net,yang2020ur2kid} that tackle both problems together and jointly try to solve them. Some efforts, as in \cite{yi2018learning,zhang2019oa-net}, train the network to learn how to reject outliers in the presence of putative matches. SuperGlue \cite{sarlin2020superglue} learns to make appropriate matches using \textit{matching descriptors} which are obtained by benefiting both original descriptors and keypoint locations via multilayer perceptron. Another approach \cite{bhowmik2020reinforced} uses reinforcement learning to optimize feature detection and description for high-level tasks like relative pose estimation. Some recent approaches \cite{brachmann2019neural,brachmann2017dsac,brachmann2019esac} try to find differentiable alternatives to RANSAC in order to estimate geometric transformation better. Note that all the aforementioned efforts provide some improvements on some parts of the classical pipeline.

The performance of the overall conventional system is dictated by the weakest sub-block in this serial pipeline during image matching. Therefore, trying to enhance the performance of each stage would be sub-optimal as the output of each block is the input to another, and every single step is prone to error in a different scenario. Hence, it can be observed that some recent works in the literature have abandoned the classical pipeline and started to propose combining many stages together in their solutions to overcome this bottleneck. In this manner, there are methods \cite{detone2016deep,nguyen2018unsupervised,rocco2017cnnformatching} that handle image matching problem with a single architecture by directly estimating the geometric transformation between two images with regression. More recently, some efforts \cite{rocco2020ncnet,rocco2020sparsencnet,zhou2020patch2pix} find correspondences between two images by benefiting from deep features with a single network instead of separately detect, describe and match them.

The representations that are obtained by off-the-shelf feature extractors are shown to be quite effective for complex tasks,  such as object detection \cite{ren2015fasterrcnn,bochkovskiy2020yolov4,lin2017retinanet} and semantic segmentation \cite{he2017maskrcnn}; however, as far as we know, currently the only method employing off-the-shelf feature extractors is D2-Net \cite{dusmanu2019d2net}. In other words, the requirement of designing new deep networks for image matching is not clearly justified by demonstrating the insufficiency of off-the-shelf feature extractors as they are not examined in detail. On the contrary, it is firmly believed that the human-level performance of these feature extractors for image classification task \cite{he2015human-level} indicates their semantic abstraction capability. Moreover, their success in object detection and semantic segmentation tasks reveals their good localization capability. If these pre-trained extractors are utilized in a way for exploiting both their semantic abstraction and good localization capabilities, they might have a chance to achieve satisfactory performance for image matching task as well. 
If these pre-trained extractors are utilized in such a way that exploits both their semantic abstraction and good localization capabilities, they might have a chance to achieve satisfactory performance for image matching task as well.

Recognition-by-components theory \cite{biederman1987recognition} advocates that our way of perception of objects are separating them into parts, which supports that humans might accomplish image matching task using objects as they are the main components of images. We have also been inspired by the Mental Rotation \cite{shepard1971mental} which states that the human brain does some initial rotations while looking for similarities between two objects.   

Our main motivation in this paper is to reveal the performance of a pre-trained off-the-shelf feature extractor on image matching problem by applying simple, but effective vision techniques, such as hierarchical search, coarse-to-fine strategy or preliminary alignment, as human visual system possibly employs.

We propose a method that is able to find high-quality correspondences between images, which does not require any training. The proposed method only needs a pre-trained network (e.g. VGG \cite{simonyan2014vgg} or ResNet \cite{he2016resnet}) to extract its features. We first leverage the invariances of the deepest layers to find matches between images at a more semantic level. Then we obtain a rough estimate of the geometric transformation between the images to apply the initial warping,  as in Mental Rotation \cite{shepard1971mental}. After warping one of the images, we perform hierarchical matching, also a well-known technique, from the deepest level of the feature map to the shallowest layer. With this coarse to fine strategy, we basically exploit the properties of all layers together, such as the powerful semantic property of the deep layers and precise localization of the shallow layers. 

Hence, in this study our contribution is threefold;

i. We show that a simple dense nearest neighbor search of the deep features at the terminal layer of a pre-trained network (such as a frozen VGG) still yields sufficient matches for obtaining a geometric transformation- and illumination-invariant matching. 

ii. We also demonstrate that after making a rough alignment through geometric transformation and warping in a preliminary step, the image matching performance of a subsequent block that mimics the previous stage in a hierarchical manner to refine positions of dense matches from the terminal layers gives state-of-the-art performance.

iii. Using off-the-shelf deep feature extractors with no training and only employing standard techniques, the proposed method forms a performance baseline for learning-based image matching methods.

\section{Related Work}

\textbf{Classical Image Matching Pipeline} consists of
feature detection, feature description, feature matching, and geometric transformation estimation stages, as mentioned before. Most of the recently developed methods propose some improvements
on individual stages of this pipeline. A prominent algorithm, namely SuperPoint \cite{detone2018superpoint}, which focuses on feature detection and description, utilizes MagicPoint \cite{detone2017magicpoint} detector, which is trained on a synthetic dataset and uses Homographic Adaptation as a self-supervision strategy to label MS-COCO \cite{lin2014mscoco} dataset. Next, this detector-descriptor network is trained jointly by partly sharing parameters.  This kind of combined training certainly improves the overall robustness. However, SuperPoint has limitations under some challenges, such as in-plane rotations. The authors in \cite{pautrat2020online} show that due to the trade-off between invariance and discrimination, we can not enforce the network to be  more rotation and illumination invariant. Another promising solution, SuperGlue \cite{sarlin2020superglue} gets the detected features and their descriptors as inputs, then uses a graph neural network to find out cross and self attentions between features. Afterwards, this method learns optimally matching the features by the help of the differentiable Sinkhorn algorithm \cite{sinkhorn1967concerning,cuturi2013sinkhorn}. SuperGlue argues that it could remarkably improve the matching performances of extracted features. However, its performance depends on the quality of feature detectors and descriptors in principle. Moreover, when the detector and descriptor couple change, it should be retrained to get the best performance. D2-Net \cite{dusmanu2019d2net}, on the other hand, uses feature maps of VGG-16 to find salient points and their descriptors. Although D2-Net can work with no training, the authors argue that with well-defined losses, it can be trained and might perform better. Since D2-Net uses only the specific layers of the feature extractor network, it can not utilize all the information encoded in the feature extractor. Moreover, it tends to poorly localize the features on account of using features at a low resolution. 

\textbf{Integrated Approaches} are also applied to the image matching problem recently. In integrated approaches, the algorithms take two images as input and directly compute the correspondences between them without externally detect, describe, and match features. As a typical example, NCNet \cite{rocco2020ncnet} takes advantage of dense CNN features extracted by ResNet-101. Next, NCNet generates the 4D correlation map, which will be processed later by a 4D CNN, namely neighbor consensus network. Finally, the algorithm outputs filtered matches between two images. NCNet algorithm is a novel idea and it works well specifically in semantic matching scenarios; however, the localization performance is not good, since NCNet works in low resolutions due to the large memory requirement. As an alternative integrated approach, Patch2Pix \cite{zhou2020patch2pix} uses NCNet as a baseline and, starting from local patches, it predicts pixel-level matches by using epipolar loss. The algorithm first extracts features by adapted ResNet-34 until the $4^{th}$ convolutional layer, detects match proposals using the last layer and finally, refines these proposals by using mid and fine-level regressors. Yet, this approach does not entirely benefit from hierarchical refinement and coarse warping stages, which are well-known procedures in vision research. Moreover, it does not use the very last layers of ResNet-34 in which more semantic information is encoded, and it requires training when the feature extractor backbone change. In their ablation study, Patch2Pix clearly shows that starting with the last layer and using the information of all layers before the last layer provides a better performance compared to using some previous layers or using all layers containing the last one. These experimental results also indicate the possible advantages of hierarchical refinement that is used in our proposed method.

\section{Method}
\label{sec:method}

\begin{figure*}[ht]
\centering
\includegraphics[width=\textwidth]{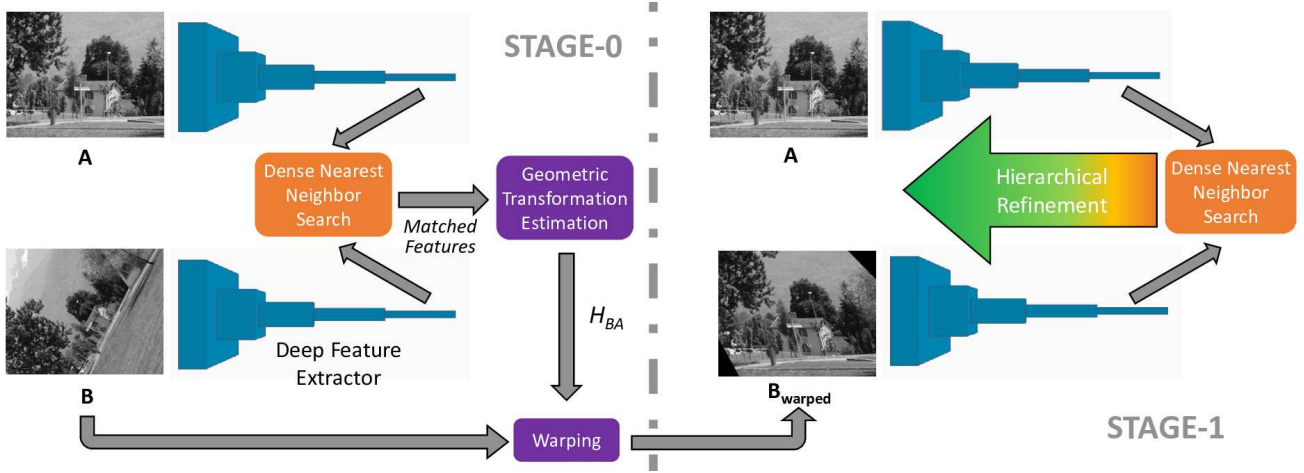}
\smallbreak
\caption{Architecture for Deep Feature Matching (DFM) Algorithm}
\label{fig:architecture}
\end{figure*}

We propose a two-stage architecture to perform matching between two images, as shown in  Figure \ref{fig:architecture}. Our key idea is to extract features with a pre-trained network, aligning images with rough geometric transformation estimation then applying a coarse to fine strategy for better localization. Opposing to the classical feature matching pipeline, rather than constraining the set of feature points by a detection step, we exploit the feature points which are successfully matched like \cite{rocco2020ncnet,zhou2020patch2pix}. In Stage-0, we perform \textit{Dense Nearest Neighbor Search} (DNNS) in low spatial resolution to obtain a rough estimation of the geometric transformation between given images. Using this estimate, we warp the second image to form the image that we use next. In Stage-1, we first perform DNNS using the feature maps of the reference image and the warped image at the last layers. Starting with these coarse matches, we refine the matches hierarchically by moving to finer resolutions at each step.  

Given an image pair, $A$ and $B$, we first perform deep feature extraction for both images using a pre-trained VGG-19 network. Then, we continue with a) Dense Nearest Neighbor Search and b) Hierarchical Refinement, which will be explained next in detail. After explaining these two basic steps, we explain the two-stage architecture and how these basic steps are employed.

\subsection{Dense Nearest Neighbor Search} 
Dense Nearest Neighbor Search (DNNS) searches for matches in dense feature maps using a mutual nearest neighbor search with a ratio test \cite{lowe2004sift} as employed in \cite{dusmanu2019d2net}. Given feature patches $F^A$ and $F^B$ extracted from images $A$ and $B$, DNNS searches $F^B$ to find the best match for each element of $F^A$. Potential matches are defined as the nearest neighbors in terms of \textit{l2} distance. For a point $p^A$ in feature map $F^A$, if the ratio of the distance to best match $p^B$ and the second-best match is below the given threshold, then the point $p^A$ is matched to $p^B$. However, the pair is accepted only if the match is mutual, i.e., $p^A$ and $p^B$ are returned as a matched pair if $p^B$ is matched to $p^A$ as well. 

We first employ DNNS to initialize matches at the terminal layer of the feature extractor. Note that, downsampled size of this layer makes them suitable for DNSS. For the VGG feature extractor, the size of the feature maps at the $5^{th}$ convolutional layer is  $W/16\times H/16$ where $W$ and $H$ are the original width and height of the image, which makes DNNS feasible. We also employ DNNS to refine the matches at finer spatial resolutions, as explained next.



\subsection{Hierarchical Refinement} 
The outputs of DNNS at the deep layers are robust to geometric transformations and illumination changes, but they are poorly localized. Directly upsampling them to the original resolution would yield poor matching performance.

As an imperfect alternative, DNNS can also be employed in former convolutional layers without any constraint. However, finer spatial resolutions of these layers would increase the computational complexity exponentially as we move towards the original resolution. Moreover, the features at deeper layers tend to carry semantic information more, which makes them more discriminative, while the features in shallower layers have smaller receptive fields and therefore tend to be more repetitive. Hence, applying DNNS at these layers brings the risk of mismatches and erroneous elimination of feature points. Patch2Pix \cite{zhou2020patch2pix} solves the latter problem by appending the features of different layers in the original resolution at the expense of computational complexity, which worsens the first problem more. However, coarse-to-fine strategy (or hierarchical search/refinement), which is a well-known technique in computer vision and practiced in many works \cite{ranjan2017spynet, sun2018pwcnet}, can be employed to solve both problems at the same time. A similar hierarchical matching strategy is also used in \cite{revaud2016deepmatching}.

Given the set of matched pairs, $P_n^{A,B}$ at the layer $n$ and feature maps $F^A_{n-1}$ and $F^B_{n-1}$ at the layer $n-1$, Hierarchical Refinement Algorithm (HRA) finds the matched points at the current layer by employing DNNS in a limited area. For each matched pair, ${p^A}$ and ${p^B}$, HRA first constructs point sets $\Omega_A$ and $\Omega_B$ which are the receptive fields of the ${p^A}$ and ${p^B}$ at layer $n-1$. For VGG-19, receptive fields can be easily obtained by upsampling the input points by 2 and including the one-pixel right and bottom neighborhood for each upsampled point. By restricting the search area as such, DNNS can be applied with none of the aforementioned problems. Therefore, given a set of matched pairs in layer $n$, for each pair, we feed the patches of feature maps $F^A$ and $F^B$ to DNNS and get the matched pairs in layer $n-1$. HRA is presented in Algorithm \ref{alg:hra}. 
\vspace*{-.1em}
\begin{algorithm}[ht]
\SetAlgoLined
 \DontPrintSemicolon
 \SetKwInOut{Input}{input}
 \SetKwInOut{Output}{output}
 \Input {$F_{n-1}^A$, $F_{n-1}^B$ \textrightarrow feature maps at layer $n-1$ \newline
  $P_n^{A,B}$ \textrightarrow set of matched pairs at layer $n$ \smallbreak} 
 \Output {$P_{n-1}^{A,B}$ \textrightarrow set of matched pairs at layer $n-1$ \medbreak}
  \SetKwFunction{FMain}{HRef}
   \SetKwProg{Fn}{Function}{:}{end}
    \Fn{\FMain{$F_{n-1}^A$, $F_{n-1}^B$, $P_n^{A,B}$}}
    {\smallbreak
     \For{$p^A$,$p^B$ \textbf{in}  $P_n^{A,B}$}{

    (1)Get the receptive fields at layer $n-1$ for feature points defined at layer $n$\newline
    $\Omega^A$ = receptive($p^A$)\newline
    $\Omega^B$ = receptive($p^B$)\smallbreak

    (2)Perform Dense Nearest Neighbor Search\newline
    $M^{A,B}$ = DNNS($F_{n-1}^A(\Omega^A)$,$F_{n-1}^B(\Omega^B)$)\smallbreak
    
    (3)Record the matched pair at layer ${n-1}$  \newline
    $P_{n-1}^{A,B}.append(transform(M^{A,B}))$\smallbreak
    }     
    
        \KwRet{$P_{n-1}^{A,B}$}
 }
 \caption{Hierarchical Refinement (HRA)}
\label{alg:hra}
\end{algorithm}

HRA changes the number of feature points in each refinement step by selecting the correct matches and rejecting the outliers. In other words, correct and robust matches are expected to survive and might even populate through hierarchical refinement, while the outliers are expected to be rejected, as illustrated in Figure \ref{fig:href}. As seen in this figure, there are some poorly localized matches in $Layer{-}5$. In $Layer{-}4$, the number of matches is increased, but still, their localization is inferior. Moving forward to $Layer{-}3$, matched points are localized much better, but there are still some erroneous matches. (Note the $2^{nd}$ and $3^{rd}$ matches from the top.) Notice that these outliers are also rejected in $Layer{-}2$. Finally, we ended up with better localized and more accurate matches in $Layer{-}1$.

\subsection{Two Stage Approach}
Given an image pair, one can extract the features by using a pre-trained feature extractor, perform DNNS at the terminal layer to initialize the matched pairs, and refine them up to the original resolution using HRA. The matching performance of this single-step solution will be presented in Section \ref{sec:experiements} as DFM(s1). However, we realized that this one-step solution fails during hierarchical refinement in case of severe geometric transformations. This result is due to the fact that features at shallower layers are not as robust as deep layers against geometric transformations. We also observed that even if the method is able to generate correct matches at deeper layers, these matches are eliminated while moving towards shallower layers. To overcome this issue, we first predict a homography matrix $H_{BA}$ using the matched set of points at the terminal layer and warp $B$ using $H_{BA}$ to obtain warped image $B_{w}$. Next, we apply the solution mentioned above by using the reference image $A$ and $B_{w}$. With such an approach, we are effectively first aligning the two images, then looking for possible matches, as proposed in \cite{rocco2017cnnformatching}. Recently, this approach is successfully applied in \cite{shen2020ransacflow,parihar2021rord,truong2020glunet}.

When VGG-19 is employed as the feature extractor, we first obtain a set of matched points by feeding $P_5^A$ and $P_5^B$ to DNNS. Using this set of matched pairs, we predict a homography matrix $H_{BA}$ and obtain warped image $B_{warped}$. We extract features one more time for the warped image. Then, we initialize a set of matched points utilizing the terminal layers' features of the warped and the reference image with DNNS. For VGG-19, each point of the pair is the parent of 4 points in the former layer. To refine the pairs, HRA applies DNNS on 2x2 feature patches iteratively until the first layer. Note that each level number of feature points might change; in other words, $P_1^A$ may have more points than $P_5^A$, or vice versa.


\section{Experiments and Evaluation}
\label{sec:experiements}
We have evaluated the proposed DFM technique on

\begin{figure}[ht]
\centering
\includegraphics[width=0.43\textwidth]{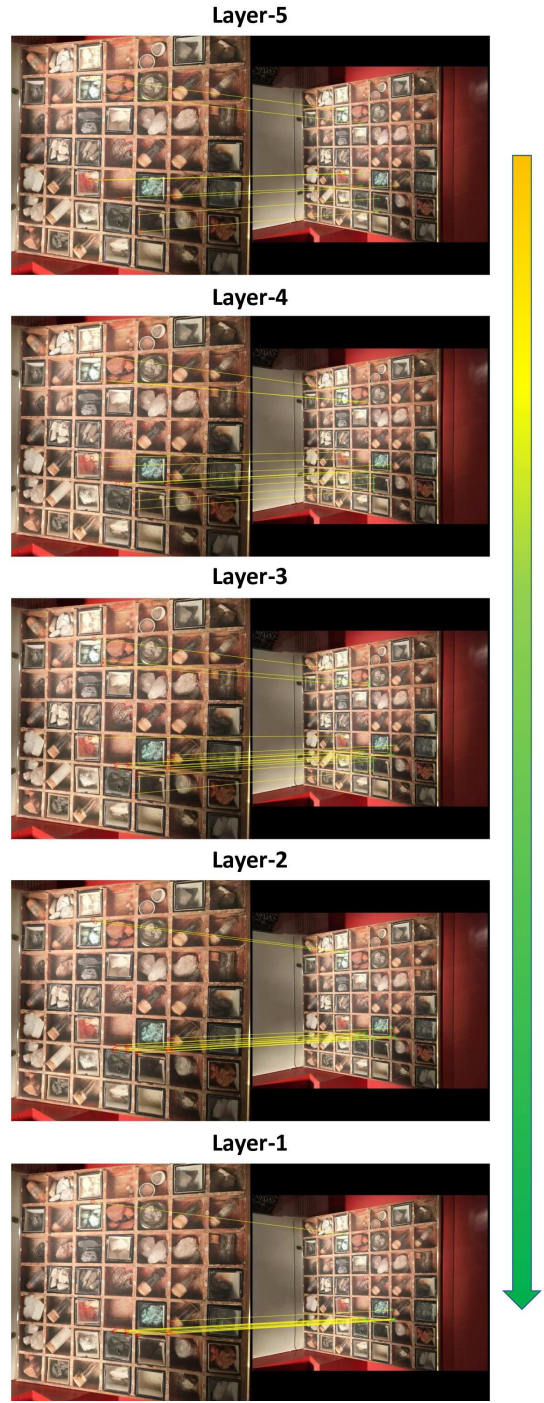}
\smallbreak
\caption{Visualization of Hierarchical Refinement Algorithm. Proposed method starts with Dense Nearest Neighbor Search at the $5^{th} layer$ of the VGG-19, then applies Hierarchical Refinement Algorithm until the $1^{st} layer$. For clarity, a limited number of matched pairs are shown.}
\label{fig:href}
\end{figure} 

\begin{figure*}[t]
  \centering
  \subfloat{\includegraphics[width=.65\textwidth]{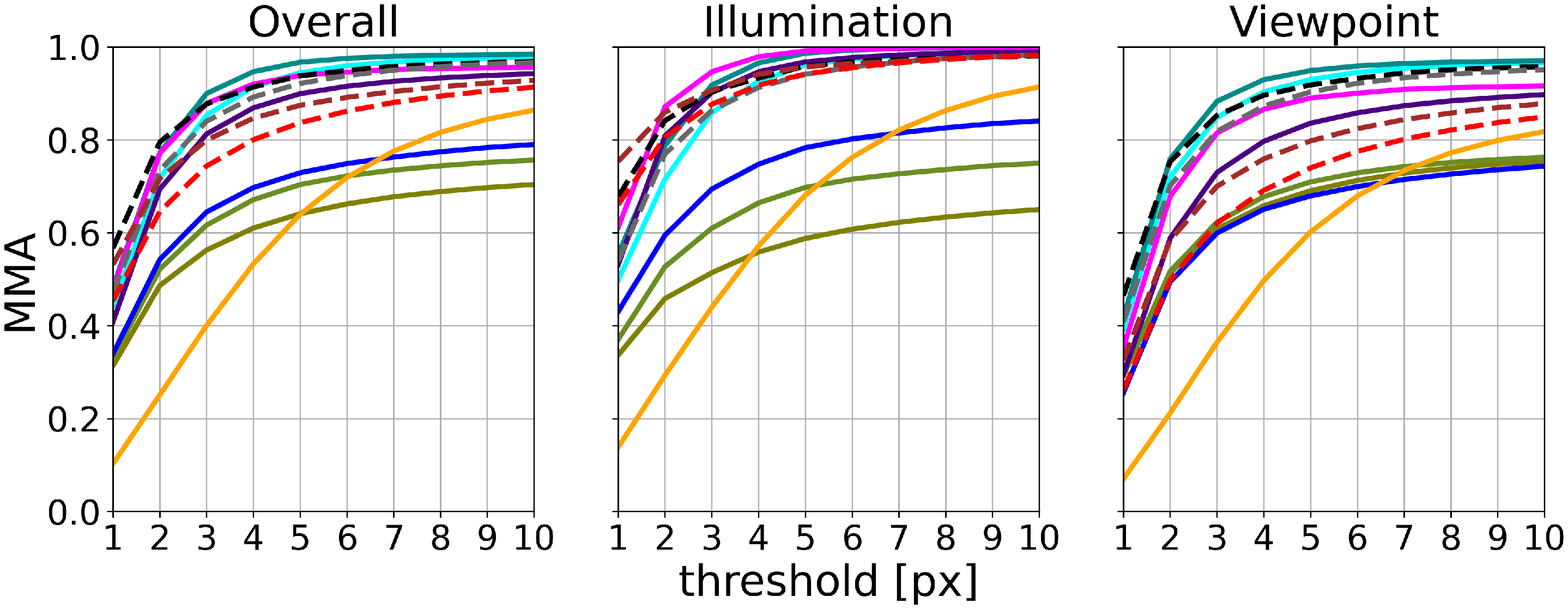}}
  \hspace*{\fill}
  \subfloat{\adjustbox{max width=0.35\textwidth}{
    \begin{tabular}{ |c@{\hskip -0.5cm}  m{5.9cm}@{\hskip -1cm}  m{3.2cm}@{\hskip -0cm} | }
    \hline
     & \centering\textbf{Method} & \textbf{\#Features\textbackslash Matches} \\ \hline
    \begin{minipage}{.03\textwidth}
    \vspace*{-0.4mm}
      \includegraphics[width=\linewidth,height=4.85cm]{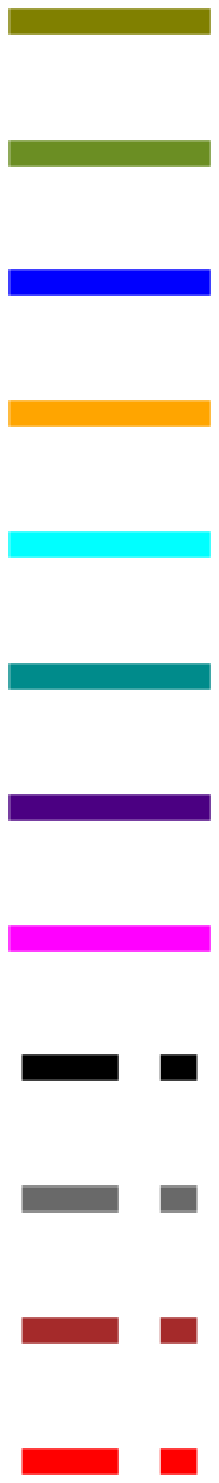}
    \end{minipage}
    &
      \begin{itemize}
      \vspace*{-1.2mm}
        \setlength\itemsep{-0.4em}
        \item[] HesAff \cite{mikolajczyk2004scale} + RootSIFT + NN
        \item[] HAN \cite{mishkin2018repeatability} + HN++ \cite{mishchuk2017hardnet} + NN
        \item[] SuperPoint \cite{detone2018superpoint} + NN
        \item[] D2Net \cite{dusmanu2019d2net} + NN
        \item[] SuperPoint + SuperGlue \cite{sarlin2020superglue} (c=0.2)
        \item[] SuperPoint + SuperGlue \cite{sarlin2020superglue} (c=0.9)
        \item[] Patch2Pix \cite{zhou2020patch2pix} (c=0.5) 
        \item[] Patch2Pix \cite{zhou2020patch2pix} (c=0.9)
        \item[] DFM(s0+s1)(r=0.6)
        \item[] DFM(s0+s1)(r=0.9)
        \item[] DFM(s1)(r=0.6)
        \item[] DFM(s1)(r=0.9)
        \vspace*{-2mm}
      \end{itemize}
    & 
      \begin{itemize}
      \vspace*{-1.2mm}
      \centering
        \setlength\itemsep{-0.4em}
        \item[] 6.7k{\textbackslash }2.8k
        \item[] 3.9k{\textbackslash }2k
        \item[] 1.6k{\textbackslash }0.9k
        \item[] 6.0k{\textbackslash }2.5k
        \item[] 0.5k
        \item[] 0.4k
        \item[] 1.1k
        \item[] 0.7k
        \item[] 4.8k
        \item[] 12.7k
        \item[] 2.9k
        \item[] 7.5k
        \vspace*{-2mm}
      \end{itemize}
    \\ \hline
  \end{tabular}}}
  \caption{Feature Matching Evaluation HPatches in terms of MMA. Variants of DFM are demonstrated as dashed lines.}
  \label{fig:mma}
\end{figure*}

\noindent HPatches \cite{balntas2017hpatches} dataset, which consists of 116 sequences in total. Each sequence has 6 images of the same scene, each captured from a different viewpoint or under different illumination. For each sequence, the dataset provides ground-truth homographies between the first image and the remaining five images. These 116 sequences are divided into two subsets; 57 sequences have significant illumination changes with the almost same viewpoint, while 59 sequences have significant viewpoint changes under similar illumination.

We have performed feature matching and homography estimation on the HPatches dataset and reported the results in the following two subsections.
\subsection{Feature Matching}
The performance of the feature matching task is measured by the pixel-wise distances between the matched features and their ground-truth projections on the pair images, as in \cite{dusmanu2019d2net}. Then, these values have been averaged for the whole dataset, and the percentage of the matches with matching error less than the selected threshold is reported for thresholds from 1 pixel to 10 pixels. This metric is known as mean matching accuracy (MMA) \cite{dusmanu2019d2net}. 

The results of various methods are illustrated in Figure \ref{fig:mma} with the number of matched features. We reported results for four different variants of the proposed DFM. We have denoted the variants that employ two-stage approach as (s0+s1) to abbreviate Stage-0 and Stage-1. Note that Stage-1 is independent of Stage-0 and can be used solely. We have represented these one-stage variants as (s1).  There are two more variants according to diversely conducted ratio tests. We abbreviate the set of ratio test thresholds $\{0.6, 0.6, 0.8, 0.9, 0.95\}$ employed in hierarchical refinement algorithm as (r=0.6) to denote the ratio threshold at the shallowest layer. For (r=0.9) the threshold set is $\{0.9, 0.9, 0.9, 0.9, 0.95\}$. For the two-stage (s0+s1) variants, we have not applied any ratio test on Stage-0. In both configurations, we have utilized conv5\_3 layers of VGG-19 on Stage-0, and we have utilized conv5\_2, conv4\_2, conv3\_2,  conv2\_2, conv1\_2 layers on Stage-1. Our implementation will be available at \url{https://github.com/ufukefe/DFM}.

As shown in Figure \ref{fig:mma}, DFM(s0+s1)(r=0.6) outperforms all other methods for the thresholds less than 3 pixels on overall results. Moreover, all variants of the DFM compete with the state-of-the-art performance as the best second or third method depending on the case even without particular training for feature matching task. In viewpoint results, the significant effect of Stage-0 can be observed clearly. 

Apart from the promising matching results, DFM has much more matches than any other algorithms. This means DFM returns denser matches, claiming that \textit{there are at least this many matches giving this MMA results}.
\subsection{Homography Estimation}
Carrying out the Feature Matching evaluation in terms of MMA is not sufficient for understanding the performance of the compared algorithms. If an algorithm finds a single correct match for each pair in the dataset, then its MMA score will be 100\% which is misleading. Researchers try to examine such a case by also reporting the average number of matches, but algorithms might still fail for specific pairs. In other words, an algorithm may have a sufficient number of matches, but all these matches may concentrate on the specific regions of the image pairs, preventing to achieve a high-quality homography estimation. For measuring this characteristic, following \cite{detone2018superpoint,zhou2020patch2pix}, we also evaluate our method in terms of homography estimation accuracy for 1, 3, and 5 pixels. For this evaluation, four corners of the reference image is projected onto the pair image with estimated and ground-truth homographies, and the average pixel distance between estimated and ground-truth projections of the four corners is obtained. Then, this average distance is thresholded to conclude whether homography estimate for a pair is correct or not, and the rate of the correctly estimated homographies are reported in Table \ref{tab:homography}.

\begin{table}[b]
\adjustbox{max width=\columnwidth}{
  \begin{tabular}{clccc}
    \toprule
     \multirow{2}{*}{\shortstack{Reported \\in}} &
      \multirow{2}{*}{Method} &
      \multicolumn{3}{c}{Homography Estimation Accuracy}\\
      \cmidrule{3-5}
    {} & {} & {$\leq$1px} & {$\leq$3px} & {$\leq$5px} \\
    \midrule
    \multirow{1}{*}{\cite{detone2018superpoint}} & SuperPoint \cite{detone2018superpoint} + NN & 0.31 & 0.68 & 0.83 \\
    
      \midrule
      \midrule
    \multirow{6}{*}{\cite{zhou2020patch2pix}} & SuperPoint \cite{detone2018superpoint} + NN & 0.46 & 0.78 & 0.85  \\
    {} & D2Net \cite{dusmanu2019d2net} + NN & 0.38 & 0.72 & 0.81 \\
    {} & R2D2 \cite{revaud2019r2d2} + NN & 0.47 & 0.78 & 0.83 \\
    {} & SuperPoint + SuperGlue \cite{sarlin2020superglue} & 0.51 & 0.83 & 0.89 \\
    {} & SparseNCNet \cite{rocco2020sparsencnet} & 0.36 & 0.66 & 0.76 \\
    {} & Patch2Pix \cite{zhou2020patch2pix} & 0.51 & 0.79 & 0.86  \\
    \midrule
    \midrule
    {} & DFM(s0+s1)(r=0.6) (boe) & 0.53 & 0.84 & 0.90 \\
    {} & DFM(s0+s1)(r=0.6) ($\mu \pm \sigma$) & 0.43$\pm$.01 & 0.73$\pm$.01 & 0.83$\pm$.01 \\
    {} & DFM(s0+s1)(r=0.6) (woe) & 0.34 & 0.62 & 0.73 \\
    \midrule
    {} & DFM(s0+s1)(r=0.9) (boe) & 0.58 & 0.89 & 0.93 \\
    {} & DFM(s0+s1)(r=0.9) ($\mu \pm \sigma$) & 0.43$\pm$.01 & 0.76$\pm$.01 & 0.87$\pm$.01 \\
    {} & DFM(s0+s1)(r=0.9) (woe) & 0.27 & 0.60 & 0.76 \\

    \bottomrule
  \end{tabular}
  }
\caption{Homography Estimation results on HPatches  \cite{balntas2017hpatches}. DFM results are presented in terms of average accuracy and its standard deviation over 10 runs. Selecting the best outcome for each image pair (boe) and the worst outcome for each image pair (woe) result in significant deviation. $r$ is the used ratio test threshold.}
\label{tab:homography}
\end{table}

As various setups had been used for this evaluation \cite{detone2018superpoint,zhou2020patch2pix,sarlin2020superglue,sun2021loftr}, it is relatively difficult to make firm conclusions from Table \ref{tab:homography}. Even if almost in every setup RANSAC is employed for homography estimation, independent setups utilized different implementations (OpenCV and pydegensac). The parameter selections of RANSAC (the number of iterations, confidence and distance thresholds) are not clearly specified in all these previous setups. Moreover, the stochastic nature of the RANSAC is completely ignored. Besides, every method has performed this test by using the dataset images at different resolutions. For example, there is a remarkable amount of discrepancy between the reported results for the SuperPoint algorithm in \cite{detone2018superpoint,zhou2020patch2pix} as shown in Table \ref{tab:homography}. 

In order to report the homography estimation performance of the proposed approach, we used M-estimator SAmple Consensus (MSAC) \cite{torr2000mlesac} algorithm by employing  \textit{estimateGeometricTransform} function of MATLAB. Parameters of the function are \textit{'Confidence'}-trust of detecting the highest number of inliers-, \textit{'MaxNumTrials'}-limit for random trials to find inliers- and \textit{'MaxDistance'}-the maximum pixel-wise distance between a point and back projection of its pair for inliers-. These parameters were selected as \textit{99.99, 5000, 3} respectively, and matched features from full-resolution images were used in our experiments. We reported the mean and variance of correctly estimated homographies to take MSAC's stochastic nature into account. As we achieve a small standard deviation in 10 repetitions, it can be concluded that the number of repetitions is enough to obtain statistically meaningful results for the overall dataset. We have also reported the maximum and minimum possible results by taking the best outcome for each pair (boe) and the worst outcome for each pair (woe) among 10 repeated tests. As shown in Table \ref{tab:homography}, \textit{boe} and \textit{woe} results differ significantly, which indicates either our MSAC configuration is not repetitive enough for individual pairs or our putative match sets is not distributed evenly over the image pairs. This approach might seem like cheating, in fact it is, and hence they are not used in comparisons; however, as the number of repeated tests is very limited, the results cannot be explained by randomness only. These results indicate that there are enough correct pairs to achieve \textit{boe} homography estimation accuracy.

As shown in Table \ref{tab:homography}, DFM(s0+s1)(r=0.9) variant is the second-best algorithm for the 5-pixel threshold in terms of average accuracy. Nevertheless, even we believe the necessity of this evaluation, we avoid making further comments due to the inconsistencies.

\section{Conclusion}
In this paper, we have proposed a method for finding point correspondences between two images by using the features extracted by deep neural networks. We show that benefiting from basic vision techniques and using the features extracted by off-the-shelf deep neural networks, state-of-the-art performance can be achieved without any special training for feature matching. We demonstrate that the features extracted by VGG network at its deepest layers are quite discriminative that a simple nearest neighbor search can be applied densely to initiate matches, decreasing the necessity of a learning-based dense feature matcher, like NCNet \cite{rocco2020ncnet}. We also present our results that these initial matches can be refined hierarchically through shallower layers.


Nonetheless, as a limitation, for initial warping we assume that the matched scenes are planar. Hence, our two-stage solution tends to perform well in planar scenes as demonstrated on HPatches dataset \cite{balntas2017hpatches} and may be useful for aerial image matching while it may have some problems in the presence of non-planar image pairs. For such scenarios, the proposed method should be either modified to solve appropriate geometric transformation instead of homography, or should be utilized with one-stage version. For the one-stage case, weakness of features at shallow layers limit the performance by preventing the method from finding matches at those layers, which we observed in the Hpatches dataset, in case of severe geometric transformations.

{\small
\bibliographystyle{ieee_fullname}
\bibliography{6-references}
}

\end{document}